# Learning Syntactic Dense Embedding with Correlation Graph for Automatic Readability Assessment

Xinying Qiu[1]   Yuan Chen[1]   Hanwu Chen[1]   Jian-Yun Nie[2*]   Yuming Shen[1*]   Dawei Lu[3]

[1]School of Information Science and Technology,
Guangdong University of Foreign Studies, China
[3]School of Liberal Arts, Renmin University of China

[2]Department of Computer Science
and Operations Research,
University of Montreal, Canada

xy.qiu@foxmail.com   nie@iro.umontreal.ca
ymshen2002@163.com   wedalu@163.com

## Abstract

Deep learning models for automatic readability assessment generally discard linguistic features traditionally used in machine learning models for the task. We propose to incorporate linguistic features into neural network models by learning syntactic dense embeddings based on linguistic features. To cope with the relationships between the features, we form a correlation graph among features and use it to learn their embeddings so that similar features will be represented by similar embeddings. Experiments with six data sets of two proficiency levels demonstrate that our proposed methodology can complement BERT-only model to achieve significantly better performances for automatic readability assessment.

## 1 Introduction

Readability is the ease with which a reader can understand a written text[1]. Predicting readability has been widely applied in education (Lennon and Burdick, 2004), book publishing (Pera and Ng, 2014), marketing (Chebat et al., 2003), newspaper readership (Pitler and Nenkova, 2008), and health information communication (Bernstam et al., 2005). Ever since the first study by Lively and Pressey in 1923, many researchers have developed various popular readability formulas including Flesch (Flesch, 1948), Fog (Gunning, 1969) and Lexile (Stenner et al., 1988). These traditional readability formulas are favored by domain applications due to their simplicity even though the formulas are mostly based on shallow features and known to lack accuracy (Bruce et al., 1981; Davison and Kantor, 1982; Graesser et al., 2004). Its strong reliance on expert knowledge is also a burden to adapt it to a new domain.

Machine learning approaches, which incorporate a broader set of morphological, lexical, syntactic, and discourse features, have shown to achieve better accuracy in readability assessment (Si and Callan, 2001; Collins-Thompson and Callan, 2005). Figure 1 (a) describes a generic machine-learning framework for Automatic Readability Assessment (ARA) where manual feature engineering is an important step to extract important linguistic features for building readability classification models.

To bypass the necessity of heavy feature engineering, deep learning strategies have been studied to automatically detect patterns or extract features related to readability (Azpiazu and Pera, 2019; Martinc et al., 2019; Mohammadi and Khasteh, 2019). Figure 1 (b) provides a generic neural network structure of deep learning approach to ARA. While neural network models take word embedding as input, they in general discard linguistic features traditionally used in machine learning models (Deutsch et al., 2020). If ever incorporated, linguistic features such as POS and morphological tags are only used to guide attention mechanism for embedding representation of the text (Azpiazu and Pera, 2019). Pre-trained models such as BERT (Devlin et al., 2019) learn dense representations of text by informing the models with semantically neighboring words, sentences, or context. Despite the attempts of recent research to assess BERT's ability to implicitly capture the structural properties of language (Goldberg, 2019; Jawahar et al., 2019; Kovaleva et al., 2019), it has been observed that BERT "tends to rely more on semantic than structural differences during the classification phase and therefore performs better on problems with distinct semantic differences between classes" (Martinc et al., 2019). There is clearly a lack of explicit consideration of syntactic (and structural) features in the current BERT-based models for ARA, which is known to be crucial. In this study, we address the problem of augmenting the ability of BERT with widely used linguistic features in ARA.

---

* Corresponding authors.

[1] https://en.wikipedia.org/wiki/Readability

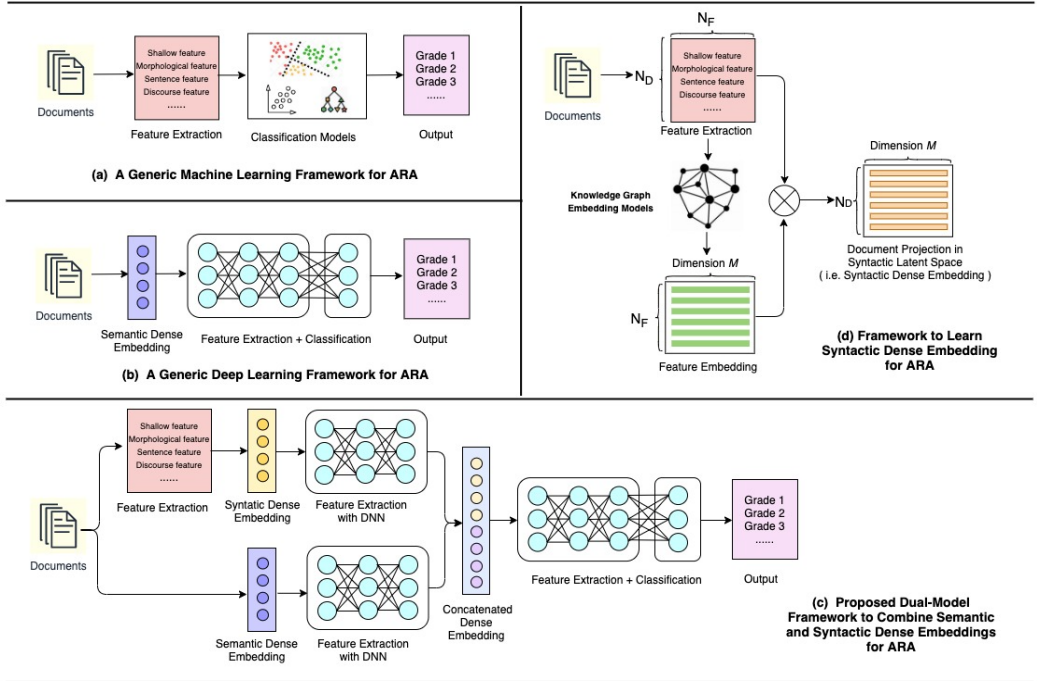

Figure 1: Proposed Dual-Model framework (c) as compared with generic machine learning (a), and generic deep learning framework (b) for Automatic Readability Assessment (ARA). Framework to learn syntactic dense embedding for ARA is provided in (d).

|   | Linguistic Feature 1 | Linguistic Feature 2 | Correlation | Latent Factors |
|---|---|---|---|---|
| 1 | Percentage of conjunctions | Average height of parse tree | Positive | Complex parse tree contains more conjunctions. |
| 2 | Average number of characters per word | Percentage of unique functional words | Negative | Length of Chinese functional word is short. |
| 3 | Number of clauses per sentence | Average number of unique idioms per sentence | Unrelated | Neutral |

Table 1. Motivating examples for constructing knowledge graph with Chinese L1 linguistic features for ARA

To best integrate with BERT, we create syntactic dense embedding as shown in Figure 1(d). An important problem we consider in this paper is the possible relationships between different features. Linguistic features defined by linguistic experts may often be related. Table 1 shows three pairs of linguistic features for Chinese readability assessment. In example one, the "percentage of conjunctions" and the "average height of parse tree" may be positively correlated because both reflect the complexity of the sentences. In the second example, "percentage of unique functional words" in a document is negatively correlated with the "average number of characters per word" for that document because Chinese functional words are usually short (i.e., one or two characters). Utilizing all these linguistic features as if they were independent may potentially hinder the classifier. We propose to consider the possible relationships among linguistic features when creating their dense embeddings with which we could complement the BERT embedding representations.

In this paper, we represent pairwise correlations between features as triplets with linguistic features as nodes and their correlations as edges. Positive correlation implies that two features behave similarly in influencing the readability level of the text and should be represented with similar embeddings. The set of triplets forms a graph (as illustrated in Figure 1(d)). We then learn the dense representations of linguistic features with graph-based models. By encoding the similarity knowledge with dense embeddings, the ARA classifier models will be better informed and gain predictive strength. Our experiments on six datasets will confirm the effectiveness of this approach.

We contribute to the research on Automatic Readability Assessment in the following directions: (1) We provide three new data sets of linguistic features for document-level readability assessment of Chinese L1, Chinese L2 and English L2 learning. (2) We verify that the correlation relationships among linguistic features could be utilized to learn syntactic dense embeddings. (3) We propose a Dual-channel neural network model (i.e., Dual-Model) to combine the syntactic dense embeddings and the BERT semantic dense embeddings for readability predictions. (4) We verify, with six data sets of Chinese and English corpora for L1 and L2 language proficiencies, that the Dual-Model can significantly improve the predictive performances of the BERT-only model. We provide our data and codes at: https://github.com/luv2Lab/linguistic-feature-embedding.

## 2 Related Work

### 2.1 Automatic Readability Assessment

Corpora for readability assessment are available for many languages. Among some of the most cited of English readability assessment are the WeeBit corpus by Vajjala and Meurers (2012, 2014) for English L1 learning and the Cambridge exam corpus by Xia et al. (2016) for English L2. For Chinese readability assessment, Sung et al. (2015) evaluated 30 linguistic features and classification models with text books in traditional Chinese. Qiu et al. (2017), Lu et al. (2019), and Zhu et al. (2019) designed features of different categories for machine learning methods for Chinese L1 and L2 readability assessment at document and sentence levels. Similar works on other languages include French (Todirascu et al., 2016), German (Hancke et al., 2012), Swedish (Pilán et al., 2016), and Japanese (Wang and Andersen, 2016). Azpiazu and Pera (2020) analyzed the most common linguistic features for six languages and evaluated multiple classifiers for cross-lingual readability assessment.

Most of the current work on applying graph-based methods or neural networks to readability assessment operate with word-level semantic embeddings. For example, Jiang et al. (2018) incorporated word-level difficulty from lexical knowledge sources into knowledge graph and trained enriched word embedding representations. Martinc et al. (2019) applied three types of neural language models at word level for unsupervised assessment. Mohammadi and Khasteh (2019) simplified the process of feature extraction with GloVe model for word embedding and reinforcement learning for English and Persian readability assessment. Azpiazu and Pera (2019) presented a multiattentive recurrent neural network model that considers raw words as input and incorporates attention mechanism with POS and morphological tags. Deutsh et al. (2020) proposed a fusion model by adding the numerical output from transformer to the linguistic features as input into SVM classifiers for readability prediction.

We notice that in previous studies, the linguistic features are mostly considered to be independent. Each of them is used as an additional one to another. However, two features can reflect the same type of linguistic phenomenon, and thus are positively correlated in influencing the readability of a text. The correlation relationships among features may help learn dense representations of linguistic features to be utilized by neural network models for better-informed predictions.

### 2.2 Feature Embedding

An important question in building neural network models is how to learn embedding representation. Feature binning has been studied to exploit the relatedness between different intervals of feature values in feature vector representation (Sil et al., 2017; Liu et al., 2016). In particular, Maddela and Xu (2018) applied smooth binning and project each numerical feature into a vector representation with multiple Gaussian radial basis functions. The embedding approach captures the nuance relationships between different intervals of feature values.

Methods similar to word embedding (Mikolov et al., 2013) have been applied to create embeddings of POS tags. Chen and Manning (2014) showed that the POS tag and arc labels exhibit semantic similarity like words and embedding can capture the similarities between POS tags or arc labels. We hypothesize that the pair-wise correlations among the linguistic features for ARA can also be used to learn embedding and we propose to use graph-based model for that purpose.

There exists a vast amount of research on graph-based embedding (Nickel et al., 2016; Wang et al., 2017; Cai et al., 2018; Ji et al., 2020). We study two methodologies in particular: Retrofitting (Faruqui et al., 2014) and TransE (Bordes et al., 2013).

The resulting similarities learned from data-driven embedding may not fully reflect the similarities one has in mind for their application (Goldberg, 2017). Retrofitting (Faruqui et al., 2014) used information from WordNet, Framenet and PPDB to improve pre-trained embedding vectors so that related words will have more similar embeddings. The method first constructs a graph $(V, E)$ where $V$ is the set of word types, and $E \subseteq V \times V$ indicates semantic relationships among pairs of words with ontology $\Omega$. Given an original embedding vector $\hat{q}_i$, a new embedding $q_i$ is learned such that it is closer to $\hat{q}_i$ and its neighbors $q_j$, $\forall j$ such that $(i,j) \in E$ and with closeness measured by Euclidean distance. The objective is to minimize $\Psi(Q)$:

$$\Psi(Q) = \sum_{i=1}^{n}\left[\alpha_i \|q_i - \hat{q}_i\|^2 + \sum_{(i,j \in E)} \beta_{ij} \|q_i - q_j\|^2\right]$$

where $\alpha$ and $\beta$ control the importance of a word embedding $q_i$ being similar to itself in the original space or to another word in the same space connected by relational information.

While Retrofitting is used to improve entity embedding in a graph, knowledge graph embedding learns representations for both the entities and their relations. TransE is a representative translational distance model where entities and relations are modeled in the same Euclidean space. Given two entity vectors $\boldsymbol{h}, \boldsymbol{t}$ and a translation vector $\boldsymbol{r}$ between them, the model requires $\boldsymbol{h} + \boldsymbol{r} \approx \boldsymbol{t}$ for the observed triple $(h, r, t)$. Hence, TransE assumes the score function

$$f_r(h, t) = \|\boldsymbol{h} + \boldsymbol{r} - \boldsymbol{t}\|_{L_1/L_2}$$

is low if $(h, r, t)$ holds, and high otherwise. To differentiate between correct and incorrect triples, TransE score difference is minimized using margin based pairwise ranking loss.

## 3 Methodology

Let $F = \{f_1, \dots, f_{N_F}\}$ (where $N_F$ is the number of features) be a linguistic feature set designed for readability assessment. Let matrix $\mathcal{D}$ be a collection of the vector representations of $N_D$ documents with $d_i \in R^{N_F}$, where $d_i = (x_1, \dots, x_{N_F})^T$, and $x_j (1 \le j \le N_F)$ is the value of feature $f_j$ in $d_i$. To construct the syntactic dense embeddings for document representation, we perform the following steps:

(1) We apply Gaussian-binning method (Maddela and Xu, 2018) to $\mathcal{D}$ such that each feature value $x_j$ of $f_j$ in $d_i$ is projected into a $k$-dimensional vector $\overrightarrow{x_j} = (y_1, \dots, y_k)^T$ where $y_n$ $(1 \le n \le k)$ is the distance of feature value $x_j$ in $d_i$ to bin $n$. We concatenate the $\overrightarrow{x_j}$ for all $d_i$ to form the initial data-driven embedding of feature $f_j$, with dimension of $M = k \times N_D$, $\forall j \in N_F$.

(2) We form a feature graph $\mathcal{G}$ using positive correlations among the $N_F$ features by setting a correlation threshold of 0.7. We preserve only the positive correlations in the graph.

(3) Let the matrix $L \in R^{M \times N_F}$ be the collection of embeddings of $f_j \in R^M$. Given a feature graph $\mathcal{G}$ and matrix $L \in R^{M \times N_F}$, we apply TransE (Bordes et al., 2013) or Retrofitting (Faruqui et al., 2014) to learn optimized feature embeddings for each feature $f_j$. Instead of random initialization, we use the data-driven embedding of $f_j \in R^M$ from Step (1) as the initial entity embedding for optimization. The syntactic latent space of $R^M$ is trained by TransE or Retrofitting respectively to encode the relationship knowledge implied by the correlations among linguistic features so that the final dense embedding of linguistic feature $f_j$ will be closer to those positively correlated with it in graph $\mathcal{G}$. We denote the matrix optimized by TransE or Retrofitting with $L_o \in R^{M \times N_F}$.

(4) To construct the syntactic dense embeddings of document representation with the embeddings of linguistic features, we perform a linear mapping to project the document feature vectors onto the syntactic latent space $R^M$. Specifically, given a feature vector of document $d_i \in R^{N_F}$, and an optimized syntactic matrix $L_o \in R^{M \times N_F}$, the projected document vector $\hat{d}_i$ in the syntactic latent space $R^M$ is defined as:

$$\hat{d}_i = L_o d_i = (l_1, \dots, l_M)^T$$

where $l_p (1 \le p \le M)$ is the projected value of the $N_F$ linguistic features of document $d_i$ at dimension $p$ of $R^M$. We name $\hat{d}_i \in R^M$ the "syntactic dense embedding".

To construct semantic dense embeddings for the documents, we learn the BERT average embedding representations following the original procedures as shown in Figure 2, where the final BERT representation is the average over all tokens. An alternative approach is to use the [CLS] token embedding to represent the text and fine-tune it for prediction. In our pilot study, we experimented rigorously with different finetuning strategies for each of the six datasets. The best finetuning results as compared with the original BERT average embeddings are reported in Appendix A. The sizes of our corpora are small ranging from 326 to 2500 as described later in Table 2. The finetuning process for BERT with 110M parameters may fit very well on training set but may not generalize well on test set. In the pilot study, we found that the overall performances of the finetuned BERT are not better than the original BERT. Therefore, we present experimentations with the original average BERT embeddings.

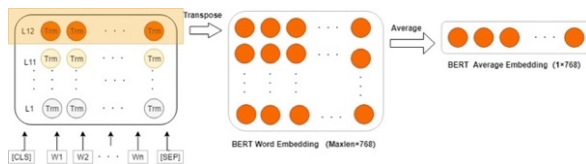

Figure 2: BERT average embedding

With the BERT dense embeddings and the syntactic dense embeddings, we propose a DNN dual channel neural network model (i.e., Dual-Model) to predict the documents' readability levels. We first feed the BERT embeddings into a four-layer network and the syntactic dense embeddings into a two-layer network. We then concatenate the outputs of the two channels into combined syntactic-semantic dense embeddings as input into another two-layer network, with MLP and SoftMax layers for readability classification. The Model architecture is provided in Figure 3.

## 4 Experiments

### 4.1 Data Sets

To evaluate our proposed models, we use six readability data sets as shown in Table 2. We create three data sets for Chinese L1 and L2 and English L2 readability assessment. The Chinese L1 data sets are textbooks for first language learning for primary school, secondary school, and high-school education from three publishers. The Chinese L2 data sets are from 5 grades of 73 textbooks that are most widely used by 7 universities in China for teaching Chinese to international students, as described in Lu et al. (2019). The ENEW data set is of 4 grades of English textbooks from New Concept English series which is one of the most widely used English L2 textbooks in China. We followed the data preparation of ENCT in Jiang et al. (2018) to prepare ENEW corpus. The raw data of Chinese L1and ENEW data sets are publicly available from their textbook websites[2,3].

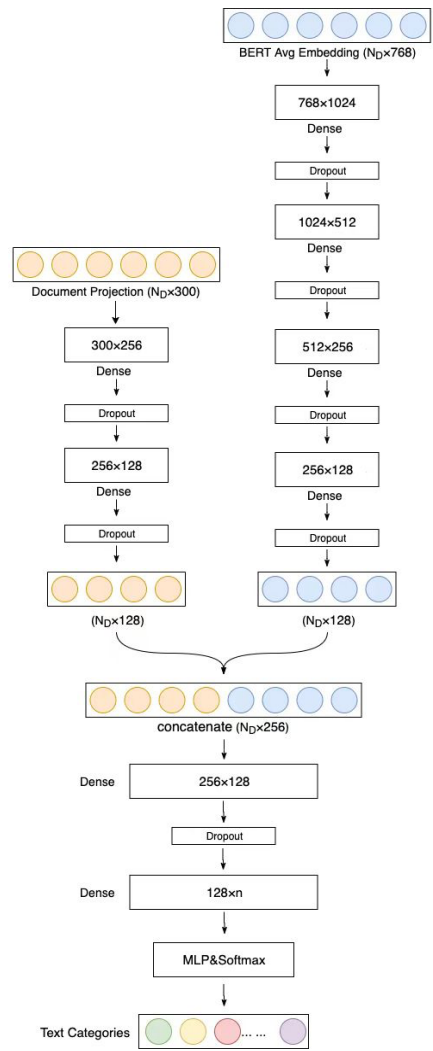

Figure 3: Dual-Model to combine syntactic and semantic dense embeddings for ARA.

In addition, we use three benchmark corpora. We obtain the WeeBit data for English L1 from the authors of Vajjala and Meurers (2012, 2014). We re-extract the text from the HTML files and discard documents that are fill-in-the-blank tests or duplicate. We take the middle set of 500 documents by document length for each class to form a 2500-document WeeBit corpus. We obtained the Cambridge Exam data set for English L2 readability assessment (Xia et al., 2016) from their website[4]. We found 5 duplicate documents in class FCE, therefore resulting in a total of 326 documents of five grade levels. We also downloaded the OneStopEnglish data set for English L2 learning from its website[5] (Vajjala and Lučić, 2018).

---

2 http://www.dzkbw.com 3 http://www.xgnyy.com

| Data Sets and Grade | 1 | 2 | 3 | 4 | 5 | 6 | 7 | 8 | 9 | 10 | Total |
|---|---|---|---|---|---|---|---|---|---|---|---|
| **Chinese L1** | 93 | 147 | 164 | 157 | 148 | 163 | 96 | 138 | 94 | 32 | 1232 |
| **Chinese L2** | 505 | 396 | 329 | 129 | 143 | | | | | | 1502 |
| **ENEW (Eng. L2)** | 72 | 96 | 60 | 48 | | | | | | | 276 |
| **WeeBit (Eng. L1)** | 500 | 500 | 500 | 500 | 500 | | | | | | 2500 |
| **OneStopEnglish (Eng. L2)** | 189 | 189 | 189 | | | | | | | | 567 |
| **Cambridge (Eng. L2)** | 64 (KET) | 60 (PET) | 66 (FCE) | 67 (CAE) | 69 (CPE) | | | | | | 326 |

Table 2: Number of documents for each grade level in each data set

Following the feature engineering methodology in previous work (Flesch, 1948; Gunning, 1969; Kincaid et al., 1975; Yang, 1970; Feng, 2010; Jiang et al., 2014; Sung et al., 2015; Qiu et al., 2017; Lu et al., 2019), we design 102 linguistic features for Chinese L1 and 111 features for Chinese L2 readability assessment. We design 33 features for English L2 referencing Vajjala and Meurers (2012). We use the feature extraction codes provided by Vajjala and Meurers (2012) to recalculate the 46 feature values for the 2500-document WeeBit corpus. We acquire the 155-feature calculation results from the OneStopEnglish corpus. We drop the features that have zero value for all documents and obtain the values of 140 features. In our pilot study with ENEW data set, we found that our 33-feature design was effective and apply these to Cambridge corpus as well. We provide linguistic feature descriptions in Appendix B.

## 4.2 Model Evaluation

According to our methodologies, we have two implementations of the Dual-channel model to combine syntactic and semantic dense embeddings for ARA: **GFE-TransE+BERT** and **GFE-Retrofit+BERT**. Both have the same network architecture as in Figure 3. The difference is that Gaussian embedding of features are used in TransE and Retrofitting respectively to learn the optimized feature embedding based on correlation graph and then produce syntactic dense embeddings of documents. We compare our methodology with the following baselines:

(1) **SVM and LR** with document feature vector $d_i \in R^{N_F}$, which are typical classification methods based on manual features.

(2) **BERT-only DNN**: This is a BERT-DNN network which has the same architecture as the right-hand side BERT channel in Figure 3. Using BERT for representation has been found effective (Martinc et al., 2019).

(3) **Raw+BERT Model**: This model concatenates the BERT DNN channel output with raw feature vectors $d_i \in R^{N_F}$ to form input into neural network for predictions. It is to verify if feature embedding is actually needed or if we could simply augment the BERT embedding with raw feature vectors for prediction.

(4) **G-Doc+BERT**: Following Maddela and Xu (2018), for each feature $x_j (j \leq 1 \leq N_F)$ in $d_i = (x_1, ..., x_{N_F})^T$, we learn the Gaussian embedding $\overrightarrow{x_j}$ and concatenate all of them into a document embedding representation. We use this syntactic dense embedding not trained by graph relations as the left-channel input in the Dual-DNN model in Figure 3 to compare with our proposed method.

For evaluation of model effectiveness, we use Accuracy and Distance-1 Adjacent Accuracy. Adjacent Accuracy means that predicting a text to be within one level distance of the true label is still considered accurate (Heilman et al., 2008). We perform 5-fold stratified cross-validation and report average Accuracy and Adjacent Accuracy. We provide the hyper parameters of neural network models and the preprocessing procedures in Appendix C, and the test of correlation thresholds in Appendix D.

---

[4] http://www.ilexir.co.uk/datasets/index.html

[5] https://zenodo.org/record/1219041

| Data Sets | Machine Learning Model | | Single-Channel Model | | | | Dual-Channel Model | | Dual-Channel with Graph-based Feature Embedding | |
|---|---|---|---|---|---|---|---|---|---|---|
| | SVM | LR | BERT-only | G-Doc-only | GFE-TransE-only | GFE-Retrofit-only | Raw +BERT Dual-Model | G-Doc +BERT Dual-Model | GFE-TransE +BERT Dual-Model | GFE-Retrofit. +BERT Dual-Model |
| Chinese L1 | 0.3498 | 0.3157 | **0.3963** | 0.3288 | 0.3734 | 0.3759 | *0.4351* | 0.3758 | **0.4732*** | **0.457*** |
| | 0.7224 | 0.6858 | **0.7946** | 0.7054 | 0.7565 | 0.7582 | *0.7972* | 0.7492 | **0.8555*** | **0.8433*** |
| Chinese L2 | 0.4447 | 0.486 | **0.6777** | 0.6032 | 0.5519 | 0.6145 | *0.6851* | 0.5979 | 0.6824 | **0.6858*** |
| | 0.8668 | 0.8995 | **0.9674** | 0.9214 | 0.8928 | 0.9294 | 0.9661 | 0.9234 | **0.9694*** | 0.9627 |
| ENEW | 0.7975 | 0.7868 | 0.8425 | **0.8515** | 0.8408 | 0.848 | *0.8441* | ***0.9494*** | 0.9094 | 0.8766 |
| | 0.9784 | 0.9675 | 1 | 1 | 0.9892 | 0.9855 | 0.9927 | 1 | 0.9927 | 0.9927 |
| WeeBit | 0.5976 | 0.6408 | **0.8348** | 0.77 | 0.66 | 0.7572 | *0.8556* | 0.8 | **0.8672*** | **0.8732*** |
| | 0.8072 | 0.8416 | **0.9868** | 0.9176 | 0.8628 | 0.908 | *0.988* | 0.952 | 0.9844 | 0.9828 |
| One Stop English | 0.6384 | 0.7301 | **0.8157** | 0.8116 | 0.7673 | 0.7795 | *0.8233* | 0.753 | **0.8501*** | **0.8661*** |
| | 0.9683 | 0.9929 | 0.9974 | 0.9982 | 0.9859 | 1 | 0.9982 | 0.9982 | 1 | 1 |
| Cambridge | 0.6501 | 0.5952 | 0.696 | 0.6993 | 0.6258 | 0.6779 | *0.7177* | *0.7208* | **0.7487*** | **0.7852*** |
| | 0.9386 | 0.9048 | **0.9755** | 0.957 | 0.9418 | 0.9325 | *0.9849* | 0.9816 | 0.9816 | 0.9785 |

Table 3. Model comparisons. BERT's performances better than machine learning, and other single-channel models are bolded. Dual-Channel Model's performances better than BERT-only model are bolded and italicized. Performances of Dual-Channel with Graph-based Feature Embedding models (i.e., our proposed methodology) better than BERT and other Dual-channel models are bolded and starred. The best performances for each data set are bolded and underlined.

## 5 Results and Analysis

We first present the comparison of BERT-only DNN model with two traditional machine learning models of SVM and Logistic Regression, and three other single channel DNN models. Table 3 shows the Accuracy and Adjacent Accuracy in the first and second row for each data set. We observe that BERT-only DNN performs the best in five out of the six data sets except for ENEW. This indicates that semantic embedding alone is very effective in ARA with neural network models which are better than traditional machine learning models with raw feature vector representations. This result is consistent with previous studies using neural network models (Martinc et al., 2019; Azpiazu and Pera, 2019).

Next, we compare BERT-only model with the Dual-channel DNN models with Raw+BERT and G-Doc+BERT. We find that augmenting BERT with raw feature value vector or document vector based on Gaussian embedding can slightly improve the performance of BERT, showing that the raw linguistic features contain additional structural information of the text that are marginally but consistently useful to the neural models for all data sets.

The performances of our proposed method are presented in the last two columns of Table 3. We observe that the two Dual-Models achieve the best performances among all 10 models in five out of six data sets (except for ENEW) and are better than the BERT-only and the other Dual-channel models. Moreover, except for Chinese L2 where the improvement is relatively smaller, the Dual-Model improvements are significant (with Student t-test at p<0.05 level) in the other four data sets of Chinese L1, WeeBit, OneStopEnglish and Cambridge. These results strongly support our earlier hypothesis that the correlations between linguistic features can provide additional useful information to learn syntactic dense embeddings that complement the semantic dense embeddings.

Comparing the last two columns of Table 3, we can observe generally similar performances in using TransE or Retrofitting on the feature graph. In theory, we impose a strict closeness constraint in Retrofitting, but let TransE learn the embedding for the correlation relation freely. The higher flexibility of TransE did not translate into better effectiveness. We speculate that the limited amount of training data may hinder our model from taking full advantage of the flexibility of TransE.

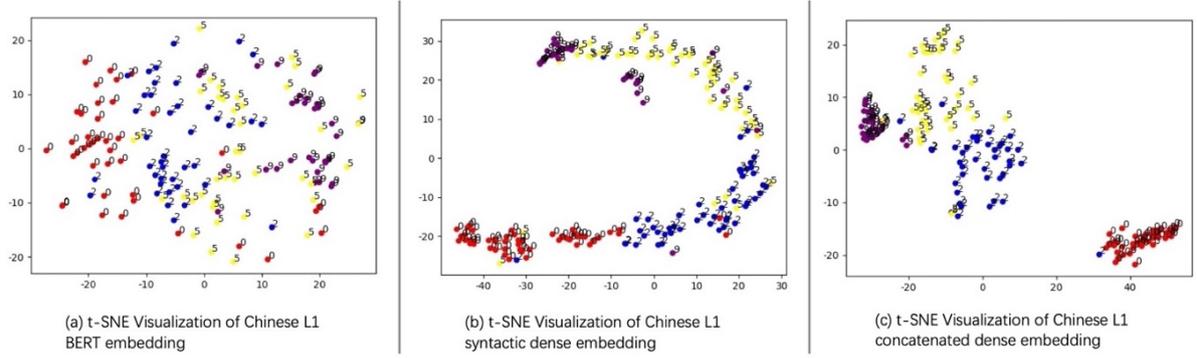

Figure 4: t-SNE visualization of semantic (a), syntactic (b), and concatenated (c) dense embeddings for Chinese L1 documents of 4 grades with grade indices of 0, 2, 5, and 9 and 40 random documents sampled for each grade.

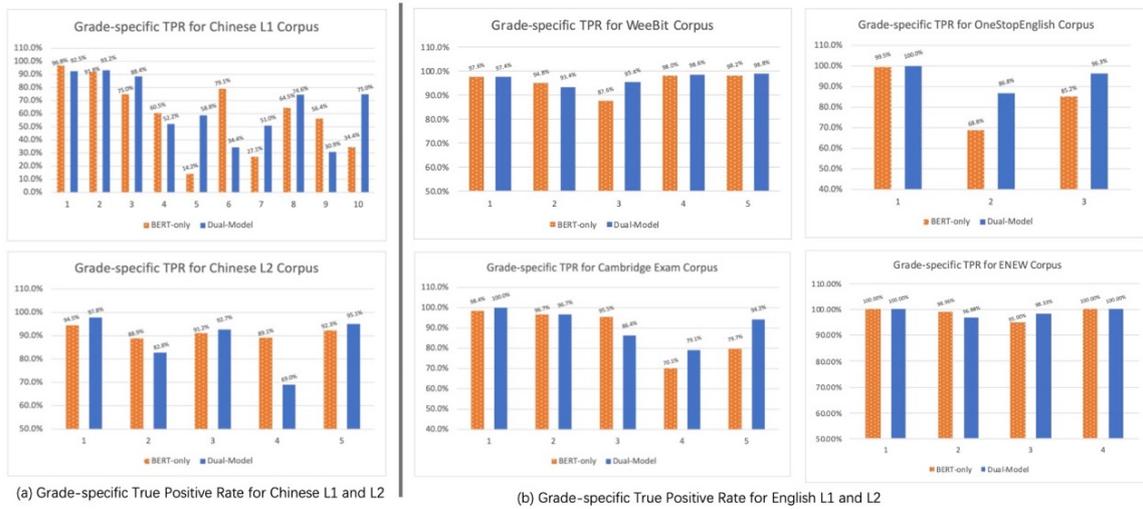

Figure 5. Compare Grade-specific True Positive Rate (TPR) of Dual-Model and BERT-only Model

Figure 4 presents a comparison of t-SNE visualization of semantic and syntactic dense embeddings, and the concatenated embedding. The figure illustrates that the concatenated embedding can produce more closely clustered data points by grade levels.

To investigate how Dual-Model improves over BERT-only model in predicting different readability levels, we present analysis of True Positive Rate (TPR) at each grade level. For each data set, we select from cross validation the best GFE-TransE+BERT model and the BERT-only model and then apply them to the whole data set. We construct confusion matrices and calculate TPR for each grade level as:

$$TPR = \frac{True\ Positive}{(True\ Positive + False\ Negative)}$$

As shown in Figure 5, for Chinese L1 we observe that the largest improvements by the Dual Model are more spread out at Grade 3, 5, 8, and 10 than for Chinese L2 which are at both ends of grade of 1 and 5. In contrast, for the four English corpora, adding syntactic dense embedding improves the BERT-only model more in the middle and the higher grade levels. We also observe from Table 3 that the improvement on Chinese L1 is more pronounced. For example, the GFE-TransE+BERT model for Chinese L1 achieved an improvement of 19.4% over BERT-only (0.4732 vs. 0.3963), while Weebit achieved an improvement of 3.88% over BERT-only (0.8672 vs 0.8348).

We may speculate that the differences in the improvement might be caused by two factors among many others: (1) how important the syntactic structure is for building the foundational knowledge in learning a certain language; and (2) how the semantic and syntactic knowledge of a certain language is organized throughout the learning process in order to lead the language learners through grasping the language.

We construct the correlation graphs with positive correlation relationship only while we observe that there exist both positive and negative correlations among linguistic features. To investigate the effectiveness of learning embedding by considering negative correlation as well, we define an additional score function for negatively correlated features used in TransE as:

$$f_r(h,t) = 1 - \|h + r - t\|_{L_1/L_2}$$

We present performance comparisons with GFE-TransE+BERT model in Table 4. We find that both models perform similarly, showing that defining positive correlation alone is sufficient in learning dense embeddings. We speculate that in feature embedding, the most important is to make similar features closer in the latent space, while repulsing negatively correlated feature embeddings away may not make a better representation of the features, which could have already been well separated in the latent space.

| Data Set | GFE-TransE+BERT Dual-Model Pos. only | GFE-TransE+BERT Dual-Model Pos. & Neg. |
|---|---|---|
| Chinese L1 | 0.4732, 0.8555 | 0.457, 0.8385 |
| Chinese L2 | 0.6824, 0.9694 | 0.6938, 0.962 |
| ENEW | 0.9094, 0.9927 | 0.9058, 0.9891 |
| WeeBit | 0.8672, 0.9844 | 0.854, 0.9792 |
| OneStopEng | 0.8501, 1 | 0.8519, 1 |
| Cambridge | 0.7487, 0.9816 | 0.7485, 0.9754 |

Table 4. Comparing performances with positive correlation-only graph and positive+negative correlation graph in GFE-TransE+BERT model

## 6 Conclusions

By combining the semantic dense embeddings and the syntactic dense embedding in a dual-channel neural network model, we propose a new methodology for readability assessment that capture both the semantic and the syntactic knowledge related to readability discrepancies. Experiments with six data sets and two proficiency levels show that our Dual-Model is better than the semantic-alone and the syntactic-alone baselines. We prove that complementing semantic dense embeddings with syntactic dense embeddings learned with correlation graph of linguistic features can produce better-informed representations for readability assessment. We will further improve our research by studying other applicable algorithms and linguistic phenomena that could benefit from learning syntactic latent space and syntactic dense embedding representations.

**Acknowledgments**

This work was supported by National Social Science Fund (Grant No. 17BGL068). We thank the anonymous reviewers for their helpful feedback and suggestions.

# Appendix A.
# BERT-finetuning Pilot Experiment Performances in Accuracy Compared with BERT original as in Paper

| Data Set | BERT-finetuning-only | BERT-only (as in paper) |
|---|---|---|
| Chinese L1 | 0.353 | 0.3963 |
| Chinese L2 | 0.5353 | 0.6777 |
| ENEW | 0.8881 | 0.8425 |
| WeeBit | 0.8016 | 0.8348 |
| OneStopEng | 0.8235 | 0.8157 |
| Cambridge | 0.6687 | 0.696 |

# Appendix B.
# Chinese L1 and L2 Linguistic Features

| Feature category | Sub-category | Features used in metrics |
|---|---|---|
| Shallow Features | Character | Common characters, stroke-counts, characters by HSK levels |
| | Words | n-gram, words by HSK levels |
| | Sentence | Sentence length |
| POS Features | | Adjective, functional words, verbs, nouns, content words, idioms, adverbs |
| Syntactic Features | Phrases | Noun phrases, verbal phrases, prepositional phrases |
| | Clauses | Punctuation-clause, dependency distance |
| | Sentences | Parse tree, dependency distance |
| Discourse Features | Entity density | Entities, named entities |
| | Coherence | Conjunctions, pronouns |

(Note: The full descriptions of the Chinese L1 and L2 features cannot be included in the paper due to space limit. Please contact the authors if needed.)

# English 33 Linguistic Features

| Category | ID | Linguistic Features |
|---|---|---|
| Lexical Features | 1 | Lexical Density (LD) |
| | 2 | Type-Token Ratio (TTR) |
| | 3 | Corrected TTR |
| | 4 | Root TTR (RTTR) |
| | 5 | Bilogarithmic TTR (LogTTR) |
| | 6 | Uber Index (Uber) |
| | 7 | Lexical Word Variation (LV) |
| | 8 | Verb Variation-1 (VV1) |
| | 9 | Squared VV1 (SVV1) |
| | 10 | Corrected VV1 (CVV1) |
| | 11 | Verb Variation 2 (VV2) |
| | 12 | Noun Variation (NV) |
| | 13 | Adjective Variation (AdjV) |
| | 14 | Adverb Variation (AdvV) |
| | 15 | Modifier Variation (ModV) |
| | 16 | Proportion of words in AWL (AWL) |
| | 17 | Avg. Num. Characters per word (NumChar) |
| | 18 | Avg. Num. Syllables per word (NumSyll) |
| Syntactic Features | 19 | mean length of a sentence |
| | 20 | average number of words per punctuation-clause |
| | 21 | number of punctuation-clauses per sentence |
| | 22 | average number of subordinate clauses per punctuation clause |
| | 23 | average number of subordinate clauses per sentence |
| 24 | average number of co-ordinate phrases per punctuation clause | |
| 25 | average number of co-ordinate phrases per sentence | |
| 26 | average number of verb phrases per punctuation clause | |
| 27 | average number of noun phrases per sentence | |
| 28 | average number of verbal phrases per sentence | |
| 29 | average number of prepositional phrases per sentence | |
| 30 | average length of noun phrases | |
| 31 | average length of verbal phrases | |
| 32 | average length of prepositional phrases | |
| 33 | average height of parse tree | |

# Appendix C.
# Neural Network Parameters and Corpus Preprocessing

| | Max Length | Batch Size | Epoch | Learning Rate |
|---|---|---|---|---|
| Chi. L1 | 512 | 4 | 60 | 0.0001 |
| Chi. L2 | 512 | 4 | 60 | 0.0001 |
| ENEW | 256 | 4 | 60 | 0.0001 |
| WeeBit | 256 | 4 | 60 | 0.0001 |
| OneStopEng. | 512 | 4 | 40 | 0.0001 |
| Cambridge | 1024 | 4 | 40 | 0.0001 |

**Corpus Preprocessing**: To calculate linguistic features, we need to first preprocess the corpus. For Chinese data set preprocessing, we use NLPIR[6] for word segmentation, LTP[7] for POS tagging and named entity recognition, and Stanford CoreNLP (Manning et al., 2014) for syntactic parsing, grammatical labeling, and clause annotation. For preprocessing of ENEW and Cambridge, we use NLTK[8] for syllable counts and Stanford CoreNLP for all other feature calculations. For WeeBit, we re-extract the documents from the HTML files and use our own procedures to reconstruct the corpus. Then we use the author's code for feature calculation (Vajjala and Meurers 2012). We use the feature values provided by OneStopEnglish directly (Vajjala and Lučić 2018).

# Appendix D.
# Test of Correlation Coefficient Thresholds

| Correlation Coefficient Threshold | Accuracy, Adjacent Accuracy |
|---|---|
| 0.3 | 0.4651,  0.8498 |
| 0.4 | 0.461,  0.8434 |
| 0.5 | 0.4635,  0.8377 |
| 0.6 | 0.461,  0.8572 |
| **0.7** | **0.4732,  0.8555** |
| 0.8 | 0.4594,  0.8385 |

**Note**: To choose an appropriate correlation coefficient threshold for constructing correlation graph, we test different thresholds on Chinese L1 corpus with GFE-TransE+BERT dual model. The above table shows that threshold 0.7 provides the best performance and therefore is used for all experiments.

---

[6] http://ictclas.nlpir.org/
[7] http://www.ltp-cloud.com/
[8] https://github.com/rlvaugh/Impractical_Python_Projects/blob/master/Chapter_8/count_syllables.py